\documentclass[11pt,a4paper]{article}
\usepackage[hyperref]{acl2020}
\usepackage{graphicx}
\usepackage{times}
\usepackage{latexsym}
\usepackage{stfloats}
\usepackage{amsmath}

\usepackage{microtype}

\aclfinalcopy % Uncomment this line for the final submission

\title{Action based Network for Conversation Question Reformulation}

\author{{Zheyu Ye,  Jiangning Liu\footnotemark[2],  Qian Yu\footnotemark[2],  Jianxun Ju} \\ Trip.com Group \\ \texttt{\{zheyuye,yuq,jxju\}@trip.com} \\ \texttt{jiangningliu.trip@gmail.com}
}
\date{}

\begin{document}
\maketitle
\begin{abstract}
Conversation question answering requires the ability to interpret a question correctly. Current models, however, are still unsatisfactory due to the difficulty of understanding the co-references and ellipsis in daily conversation. Even though generative approaches achieved remarkable progress, they are still trapped by semantic incompleteness. This paper presents an action-based approach to recover the complete expression of the question. Specifically, we first locate the positions of co-reference or ellipsis in the question while assigning the corresponding action to each candidate span. We then look for matching phrases related to the candidate clues in the conversation context. Finally, according to the predicted action, we decide whether to replace the co-reference or supplement the ellipsis with the matched information. We demonstrate the effectiveness of our method on both English and Chinese utterance rewrite tasks, improving the state-of-the-art EM (exact match) by 3.9\% and ROUGE-L by 1.0\% respectively on the Restoration-200K dataset.
\end{abstract}

\renewcommand{\thefootnote}{\fnsymbol{footnote}}
\footnotetext[2]{Work performed while Jiangning Liu and Qian Yu worked at Trip.com Group}
\renewcommand*{\thefootnote}{\arabic{footnote}}

\section{Introduction}
Reading comprehension, which aims to answer questions about a document, has made dramatic progress in recent years \citep{hermann2015teaching, joshi-etal-2017-triviaqa, rajpurkar-etal-2018-know}. Conversation question answering \citep{reddy-etal-2019-coqa, choi-etal-2018-quac} extends reading comprehension to multi-round, in which each question needs to be understood in the conversation context. Incomplete sentences are particularly tricky to deal with under the existing framework. The incompleteness of a sentence can be reflected in: 1) Co-reference, which makes the sentence vague and unclear, increases the difficulty of understanding a single sentence in the absence of context. 2) Ellipsis leads to missing sentence elements and imperfect context information. For example, the question “What year did he graduate?” cannot be understood without knowing what “he” and ellipsis behind “graduate” refer to. Solving these two problems has also triggered extensive research and discussion in the academic circle, and two sub-tasks, Co-reference Resolution and Information Completion have been derived. The above two sub-tasks can be collectively referred to as Incomplete Utterance Rewriting (IUR) or Conversation Question Reformulation (CQR). CQR, which demands a model to rewrite a context-dependent into a self-contained question with the same answer given the context of a conversation history \citep{elgohary-etal-2019-unpack}, aims to resolve the co-referential and omitted ambiguities to reconstruct the user’s original intent in a conversation accurately.

Many recent research works attempt to construct the rewrite through auto-regressive generating the target sequence with conversational history. \citealp{lin2020conversational} fine-tune the T5-base model \citep{raffel-2019-exploring}, consisting of an encoder–decoder transformer to rewrite the current question.  \citealp{vakulenko2020question} adopt a unidirectional Transformer decoder initialized with the weights of the pre-trained medium-sized GPT2 model \cite{radford2019language} to rewrite the original question. Besides, they further increase the capacity of their generative model by learning to combine several individual distributions and then produced as a weighted sum of the intermediate distributions to obtain the next token. Existing works resorted to auto-regressive mechanisms have made considerable achievements. However, auto-regressive mechanisms face three obstacles that the generative approach cannot avoid: repetitive phrase, meaningless information and semantic incompleteness. 

\begin{figure}[htbp]
    \centering
    \includegraphics[width=\linewidth]{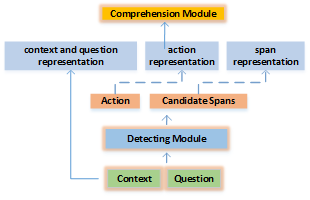}
    \caption{Overall Architecture of proposed ActNet, consisting of the encoder component, the detecting component and the comprehension component.}
    \label{fig:model-arch}
\end{figure}

\begin{figure}[htbp]
    \centering
    \includegraphics[width=\linewidth]{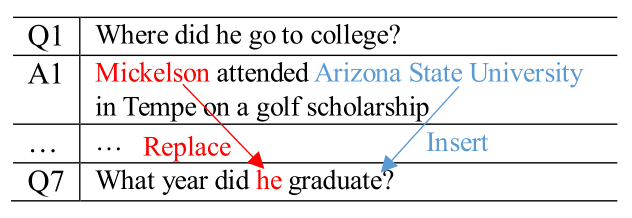}
    \caption{An example of how the action-based network works. Red means co-reference replacement and blue means omission supplement.}
    \label{fig:fig1}
\end{figure}

To address these issues, we propose an action based network (ActNet), which is composed of three modules: 1) the encoder component to encode the input text; 2) the detecting component to detect the positions of co-reference and ellipsis in the current question as well as the corresponding actions; 3) the comprehension component to find the related co-referential or omitted information from context history as illustrated in Figure \ref{fig:model-arch}. The actions consist of “insert” and “replace”, see Figure \ref{fig:fig1} for an example. Our proposed method first locate “he” and “graduate” as well as the related actions, and then find out the matched phrases “Mickelson” and “Arizona State University”. Lastly, the model replaces “he” with “Mickelson” and inserts the phrase “Arizona State University” into the position behind “graduate”.

In summary, the key contributions are 2-fold:
\begin{itemize}
    \item This paper proposes a simple but effective mechanism to recover the complete statement of the conversation question. This method can retain the original structure to the greatest extent and supplement or replace the necessary information gracefully.
    \item We demonstrate our method on both Chinese and English utterance rewrite tasks respectively. On the Restoration-200K(Chinese) dataset, ActNet outperforms the state-of-the-art rewrite model, improving exact match accuracy from 49.3\% to 53.2\%. On the CANARD(English) dataset, ActNet improves the state-of-the-art ROUGE-L score from 74.9\% to 77.6\%.
\end{itemize}

\section{Related  Work}
\paragraph{Co-reference Resolution} The co-reference resolution aims to link an antecedent for each possible mention. Traditional methods usually use pipeline structure which first detects all pronouns and entities, and then adopt cluster analysis \citep{haghighi-klein-2009-simple, lee-etal-2011-stanfords, durrani-etal-2013-markov}. Those works rely heavily on sophisticated and fine-grained features. Those works \citep{lee-etal-2017-end, kantor-globerson-2019-coreference} solved the problem in an end-to-end fashion by jointly detecting mentions and predicting co-references. \citealp{lee-etal-2017-end} introduced the first end-to-end co-reference resolution model and showed that it significantly works. However, it required computing the scores for all possible spans, which is computationally inefficient. More recently, \citealp{wu-etal-2020-corefqa} formulated co-reference resolution as query-based span prediction.

\paragraph{Text Generation} Text generation has been widely used in various natural language processing tasks. In machine translation, people use it to improve the output of the sequence-to-sequence model \citep{niehues2016pretranslation, junczys-dowmunt-grundkiewicz-2017-exploration, grangier2017quickedit, gu2017search}. In the text summary, rewriting the retrieved candidate text can provide a more accurate and complete summary \citep{see2017point, chen-bansal-2018-fast, cao-etal-2018-retrieve}. In dialogue modelling,  \citealt{weston-etal-2018-retrieve} employed it to rewrite the retrieved output.  The dominant solution to generative tasks in previous work is a collaboration between the recurrent neural network and the attention mechanism.  \citealp{gu2016incorporating} drew on the human behaviour of repeating, usually noun entities and long phrases, and proposed a copy mechanism that integrates the generation mode and copy mode. In \citeyear{see2017point}, Pointer-Generator \citep{see2017point} was first adopted into text summary with a generator to preserve its generative capabilities and a Pointers \citep{vinyals2015pointer} to copy out-of-vocabulary words from the source text to ensure repetition of crucial information.

\paragraph{Sentence Rewriting}
Sentence rewriting plays an important role in reading comprehension \cite{abujabal-2018-never} and dialogue \citep{su-etal-2019-improving, vakulenko2020question, rastogi-etal-2019-scaling, quan2019gecor, zhang2019filling}. It is proved that sentence rewrite is very helpful in understanding user intent and achieve excellent performance for task-oriented domain or open domain conversational question answering \citep{vakulenko2020question}. In the scenario of multi-talk dialogue, it is usually presented with two sub-tasks in the form of Conversation Question Reformulation. In previous work, the main idea to solve this problem is to treat it as a text generation task. with copy mechanism\citep{gu2016incorporating, su-etal-2019-improving}. Recently, \citealp{aralikatte2021ellipsis} transforms Co-reference Resolution and Information Completion problems into Question Answering form. Semantic Role Labeling (SRL) was introduced to identify predicate-argument structures of sentences, so as to capture semantic information like "who did what to whom" as priors \citep{xu2020semantic}. In addition, \citealp{liu2020incomplete} considers this task as a Semantic Segmentation task, and constructs a word-level edit matrix containing three different editing types: None, Substitute and Insert. Another con-current work \citep{hao2020robust} proposes a sequence-tagging based model to tackle the job of Sentence Rewriting. And to address the common problem, lack of fluency, of the text-generating task, \citealp{hao2020robust} inject a loss signal under a reinforcement framework.

\section{Methodology}
In this section, we will introduce our model in detail. Formally, given an utterance sequence (i.e., the conversation history) $U = \{u_1, \cdots, u_j, \cdots, u_{i-1} \}$ each could be a pair of question $Q_j$ and answer $A_j$ at the j-th turn. We unfold all tokens in $U$ into $\{w_t\}^m_{t=1}$. Symbol $m$ is the number of tokens in the conversation history. The task aims to reformulate the question $Q_i = \{q_k\}^n_{k=1}$ to the rewrite $R_i = \{r_l\}^z_{l=1}$ that contains complete intention, which can be inferenced from conversation history $U$, where symbol $n$ and $z$ refer to the number of tokens in question $Q_i$ and rewrite $R_i$ respectively.

First of all, we apply the \textbf{detecting module} to locate co-reference or ellipsis positions $\{s_k, \cdots, s_k + d_k \}$ in the question $Q_i$, where $s_k$ indicates the start position of co-reference or ellipsis and $d_k$ indicates the length of tokens in it. Note that the token is used to represent the candidate span of ellipsis when there is an ellipsis after a token. As illustrated in Figure \ref{fig:fig1}, the token “graduate” represents the candidate of ellipsis. Then we use the \textbf{comprehension module} to search for the start and end position $s_t, e_t$ of the phrase matched to the candidate from the conversation context given the representation  $h_{s_k}$ of start token $q_{s_k} $ of co-reference or ellipsis. Besides, the detecting module is also used to recognize which action $a\in \{ \text{insert, replace} \}$ the pair of phrases $\left( \{{w_t}^{s_t+d_t}_{t=s_t}\}\{{q_k}^{s_k + d_k}_{k=s_k}\} \right)$belongs to.

\subsection{Encoder}
Language model pre-training has been shown to be effective for improving many natural language processing tasks \cite{devlin-etal-2019-bert, yang2019xlnet}. These include sentence-level tasks such as natural language inference and token-level tasks such as named entity recognition and question answering. We adopt BERT-base \cite{devlin-etal-2019-bert} to map natural language text to featured representation. Specifically, we concatenate the question $Q_i$ and its context $U$ by adding special
separator tokens $<$SEP$>$ between them, hoping model can well understand the relationship between contexts. Here, we omit the subscript of $Q_i$ for simplicity. We formalize the encoding layer into the following formula:
\begin{equation}
    H^u, H^q = \text{BERT}([U;Q])
\end{equation}
where $H^u = \{h_t\}^m_{t=1}$ refers to conversation history representation, $H^q=\{h_k\}_{k=1}^n$ refers to question representation, $[\cdot;\cdot]$ means concatenating operation.

\subsection{Detecting Module}
\begin{figure}[htbp]
    \centering
    \includegraphics[width=\linewidth]{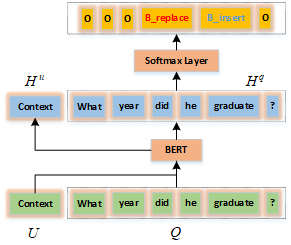}
    \caption{Architecture of encoder and detecting module. The green block is the text input, and the blue block is the representation output of that. The encoder uses a pre-trained language model BERT-base to encode the text. The question representations are injected into a softmax layer equipped with a greedy search to generate action tags.}
    \label{fig:fig2}
\end{figure}
Considering there may be several places that are co-referred or omitted in an utterance, we adopt the sequence tagging mechanism to locate the positions of them in the question. Following these works \citep{Li_2020, wu-etal-2020-corefqa}, we generate a BIO tag for each token since a co-reference span can have multiple tokens. BI tags are used to mark the beginning and inside of a span respectively. Tag O means there is no need to do anything in this place. To further distinguish each span’s characteristic, as illustrated in Figure \ref{fig:fig2}, we split tag B down into more specific cases: $B_{\text{insert}}$ to represent insert situation and $B_{\text{replace}}$ to represent replacement situation.

In addition, the tag $B_{\text{insert}}$ or $B_{\text{replace}}$ has another significance to determine the specific operation to be performed between the text segment discovered in this module and the matched text segment obtained by the comprehension module. 

We employ a linear layer followed by a softmax activation to the final encoding output of the question. The probability of assigning a tag is computed as follows:
\begin{equation}
    p_k^{\text{tag}} \propto \exp (W_d h_k  + b_d)
\end{equation}
where $W_d$ and $b_d$ are trainable parameters.

In the inference stage, we employ the greedy search to decode the tag sequence $\{\text{tag}_k\}_{k=1}^n$ of the question. The spans that started with tag B are the candidates of co-reference and ellipsis, that is $s_k=k$ if $\{\text{tag}\}_k \in B$. Note that we only consider these spans whose beginning token takes the B tag and the rest of it taking the I tag. 

\subsection{Comprehension Module}
In this module, we adopt the machine reading comprehension (MRC) mechanism \citep{hermann2015teaching, joshi-etal-2017-triviaqa, rajpurkar-etal-2018-know} to search for the information that is co-referred or omitted in the question from conversation history. For the extracted MRC tasks, a query $\hat{Q}$ and a context $ \overline{C}$ are given. The goal is to predict an answer $ \overline{A}$, which is constrained as a segment of text in the context $\overline{C}$, and neural networks are designed to model the probability distribution $p(\overline{A_c} | \overline{C}, \overline{Q_c})$. In this scenario, we define each candidate span in the question as a query $\overline{Q_c})$ and the conversation  history as context $\overline{C}$. Our goal is maximum likelihood estimation $\prod_{c=1}^{V}p(\overline{A_c} | \overline{C}, \overline{Q_c})$, where V means the number of candidate spans. 

\begin{figure}[htbp]
    \centering
    \includegraphics[width=\linewidth]{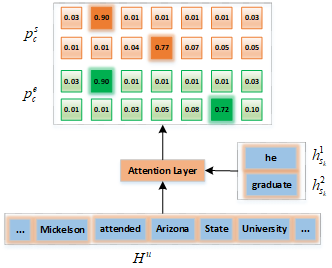}
    \caption{Architecture of comprehension module. The brown block is the probability of the start token of the answer span and the green block is the probability of the end token. The context representation and the candidate spans representation of the question is increased to an attention layer to generate answer span probability.}
    \label{fig:fig3}
\end{figure}

For simplicity, we only use the representation of the start token $q_{s_k}$ of the span as the whole semantics of the query. As illustrated in Figure \ref{fig:fig3}, we introduce a simple attention  neural network\citep{bahdanau2014neural} that produce two probability distributions over all the tokens of the given context separately for the start and end position $s_t$,$e_t$ of the answer span:
\begin{equation}
    p^{s}_{c,t} \propto \exp(v^T_s\tanh(W_s[h_t; h^c_{s_k}] + b_s))
\end{equation}
\begin{equation}
    p^{e}_{c,t} \propto \exp(v^T_e\tanh(W_e[h_t; h^c_{s_k}] + b_e)) 
\end{equation}
where $p_{c,t}^s$ refers to the probability that the t-th token in the context is marked as the starting position regarding the c-th query,$p_{c,t}^e$ refers to the probability of the end token of that.

\paragraph{Training:} We regard question rewriting as a multi-task problem, consisting of a sequence labelling task and a machine reading comprehension task. For the sequence labelling task, the training objective is defined as cross entropy loss for each token of the question: 
\begin{equation}
    L_\text{seq} = -\frac{1}{N}\sum_{i}^N\sum_k^n \log (p^{i, \text{tag}}_{y_i, k})
\end{equation}
where $y_{i,k}$ is the ground-truth tag of the k-th token of the i-th question. N is the number of examples.
The training objective of answer span prediction is defined as cross entropy loss for the start and end predictions:

\begin{equation}
    L_\text{span} = -\frac{1}{N}\sum_{i}^N\sum_V^c \log (p^{i,s}_{y_{i,c}^s})  + \log (p^{i, e}_{y_{i,c}^e}) 
\end{equation}
where $y_{i,c}^s$ and $y_{i,c}^e$ are the ground-truth start and end positions of c-th query respectively, regarding  i-th example. N is the number of examples.

During training, the joint loss function for CQR is the weighted sum of the sequence labelling loss and span loss:
\begin{equation}
    L = \alpha_1 L_\text{seq} + \alpha_2 L_\text{span}
\end{equation}
where $\alpha_1$ and $\alpha_2$ are hyperparameters.
\paragraph{Prediction:} 
During inference, we use the tag of the start token of the candidate query as an action to be performed, that is, tag $B_{\text{insert}}$ corresponds to action insert and tag $B_{\text{replace}}$ corresponds to action replace. We use the searched answer from context to replace the corresponding co-reference in the question when performing action replace and insert the searched answer after the related word when performing action insert. In the next step, we correctly copy the original question when there is no candidate query in the question.

\section{Experiments}
In this module, we briefly introduce the dataset and training data construction procedure and give the competitive experimental results on metric BLEU \citep{papineni-etal-2002-bleu}, ROUGE \citep{lin-2004-rouge} and exact match (EM) score.

\subsection{Datasets}
To evaluate the capability of our method in reformulating conversation questions, we conduct experiments on the following datasets:

\paragraph{CANARD Dataset:} An existing large open-domain dataset for Conversation Question Reformulation introduced by \citealp{elgohary-etal-2019-unpack}. Each sample in the CANARD dataset consists of an original question from the QuAC dataset \citep{choi-etal-2018-quac}, conversation history, and the corresponding rewrote question by human annotators. 

\paragraph{REWRITE:} A Chinese dataset provided by \cite{su-etal-2019-improving}. They crawled 200k candidate multi-turn conversational data from several popular Chinese social media platforms. The annotator writes the question based on the previous round of conversation and the question itself. Eventually, 40k high-quality samples with a balanced ratio of positive and negative examples, both of which has half of the data, are produced. In this corpus, positive examples contain ellipsis or reference and sentences in negative examples have complete semantic expression and need no more revision.  However, only the positive samples were published due to the company’s legal policy reasons.

\paragraph{Restoration-200K:} A large resource dataset released by \citealp{pan-etal-2019-improving}. The original corpus of this dataset was extracted from \textit{Douban Group} (A Chinese Social networking platform). As the name suggests, the dataset is 200K in size, with six utterances in each sample. The five-member annotation team spent six months completing the following two data annotation tasks: 1. A 1/0 value is consistent with the positive/negative definition in REWRITE, indicating whether the sixth utterance omits concepts or entities that appear in the first five utterances. 2. For positive examples, incomplete sentences are manually rewritten using words seen in the original text to reduce the word complexity. In addition, the annotator can use a small pre-set of commonly used words to ensure sentence fluency. In the end, they excluded about 4.8\% of the sample to produce a low-complexity dataset.

\paragraph{Dataset construction:} In order to construct the dataset that is consistent with our experimental settings, we apply SequenceMatcher\footnote{\url{https://docs.python.org/3/library/difflib.html}} method to the original question and its rewrite to find the different continuous text blocks between them. The sequence label corresponding to each text block in the original question is defined as follows: 1) the start token is tagged as $B_\text{replace}$, and the left tokens are tagged as I if necessary for the text block that is a non-empty string; 2) the token before the marked position is tagged as $B_\text{insert}$ if the text block is an empty string; 3) all tokens in original question are tagged as O if the original question is as same as its rewrite. Note that the text blocks in the original question are also queries for the comprehension module. 

We define as the answer the corresponding text block of the rewrite regarding the related query. Similarly, we use the SequenceMatcher method to search for the text segment identical to the answer from conversation history, and the first matched segment is the ground truth. Note that the sample is invalid if its answer cannot be found in the conversation history, and we will not consider putting it into the training dataset.
 
\paragraph{Data augmentation:} To collect more training samples, we also adopt the strategy for English corpus: firstly deleting some stop words, such as “for, of, to, in” etc, and then taking the same approach to generate some supplementary training samples. We found that the extra data are constructive to model. For details of stop words, please refer to Appendix \ref{sec:appendix}. Statistics of CANARD, REWRITE and Restoration-200K, are presented in Table \ref{tab:tab1}. 

\begin{table}[ht]
\begin{tabular}{cccc}
\hline
                & Train       & Dev  & Test \\ \hline
CANARD          & 11860+16310 & 3418 & 5571 \\
REWRITE         & 16081       &      & 2000 \\ 
Restoration-200K& 194k        & 5k   & 5k  \\\hline
\end{tabular}
\caption{Statistics of the datasets used in our work. The train data of CANARD contains an additional 16310 augment data.}
\label{tab:tab1}
\end{table}

\subsection{Evaluation Metrics}
Following previous works, we evaluate generation quality by leveraging three metrics: BLEU, ROUGE, and the exact match score (EM, the percentage of decoded sequences that exactly match the human references). However, our approach bypasses the idea of generation. Both BLEU and ROUGE are based on n-gram to measure the similarity between the summary of systems and humans. Specifically, BLEU measures precision: how many the machine-generated summary's words (and/or n-grams) appeared in the human reference summary. On the contrary, ROUGE measures recall: how many the human reference summary's words (and/or n-grams) appeared in the machine-generated summary.

\subsection{Baseline Models}
To verify the high efficiency of the action-based network, we compare ActNet with these baselines taking BLEU and ROUGE as the measurement, respectively. The comparison models include: 

\textbf{a) backbone models:} LSTM, a bidirectional LSTM sequence-to-sequence model with attention; GPT-2 \citep{radford2019language}, a pre-trained decoder-only transformer; BERT \citep{devlin-etal-2019-bert}, a pre-trained encoder-only transformer; UniLM \citep{dong2019unified}, where the model architecture is as same as BERT; T5 \citep{raffel-2019-exploring}, an encoder–decoder transformer.

\textbf{b) Well-designed network architecture: } T-Ptr-Gen, a transformer-based Point Generator Network \citep{see2017point}; T-Ptr-Net, a transformer-based copy network \citep{vinyals2015pointer}; T-Ptr-$\lambda$ \citep{su-etal-2019-improving}, a transformer-based copy network, where the decoder computes the probability $\lambda$ at each step to decide whether to copy from the context or the original question; TRANS-PG+BERT is a baseline model, provided by \citealp{hao2020robust}, whose structure contains a BERT encoder and a Transformer decoder that works based on a copy mechanism; The pick-and-combine (PAC, \citealp{pan-etal-2019-improving}) model consists of two steps, that is, identifying the ellipsis words in the previous sentence, and then restoring the original sentence through the identified ellipsis words; BERT+SRL\citep{xu2020semantic} uses a strong decoder with Semantic Role Labeling (SRL) as priors; RUN\citep{liu2020incomplete} consists of three components: context layer, encoding layer and segmentation layer. The original context layer is a bidirectional LSTM model. After replacing BERT as the feature extractor of the context layer, a stronger representation ability can be obtained, and this stronger network structure is RUN + BERT; Proposed by \citealp{hao2020robust}, RAST is a sequence-tagging-based model along with two reinforcement versions RAST+RL-BLEU and RAST+RL-GPT.

\subsection{Experimental Settings}
Our encoder module uses BERT-base model as the backbone, which is implemented by Tensorflow\footnote{\url{https://github.com/google-research/bert}}. We use Adam \citep{kingma2014adam} as the optimizer to update parameters and set the initial learning rate to 2e-5 with a warm-up rate of 0.1. The batch size is 24 in the training period. The number of train epochs is set to 3. The hyperparameters $\alpha_1$ and $\alpha_2$ are set to 5 and 3 respectively in this work.
\subsection{Results}
First, we do some preliminary experiments of several backbone models on CANARD Dataset as shown in Table \ref{tab:backbone}. The ActNet, fine-tuned on BERT-base, is significantly better than the LSTM and GPT-2-medium, which improves the score by 13.56\% and 2.64\%, respectively. Compared with a deeper model with a large number of parameters, like BERT-large, UniLM-large and T5-base, Actnet is not far behind, which strongly proves the superiority of our method. We suggest that the ActNet still has room for improvement when applying a pre-trained language model with better generalization regarding the comparison results (between BERT-large, UniLM-large and T5-base). 

\begin{table}[htbp]
\centering
\begin{tabular}{cccc}
                & \multicolumn{2}{l}{CANARD}  \\ \hline
                & Dev          & Test             \\ \hline
Human $\dagger$ & \multicolumn{2}{c}{59.92}         \\
Raw             & 33.84        & 36.25    \\ \hline
LSTM            & 43.68 &39.15 \\
GPT-2-medium    & 52.63 &50.07 \\
BERT-large      & 55.34 &54.34 \\
UniLM-large     & 57.39 &55.92\\
T5-base         & 59.13 &58.08 \\ \hline
ActNet + BERT   & ~ &52.71 \\ \hline
\end{tabular}
\caption{The BLEU-4 scores of different backbone models on CANARD dataset. $\dagger$: The human score computed from a small validation subset is quoted from \citealp{elgohary-etal-2019-unpack}}
\label{tab:backbone}
\end{table}

\begin{table*}[!b]
    \centering
    \begin{tabular}{lccccccc}
    \hline
        ~ & BLEU-1 & BLEU-2 & BLEU-4 & ROUGE-1 & ROUGE-2 & ROUGE-L \\ \hline
        Copy $\dagger$ & 52.4 & 46.7 & 37.8 & 72.7 & 54.9 & 68.5 \\ 
        Pronoun Sub $\dagger$ & 60.4 & 55.3 & 47.4 & 73.1 & 63.7 & 73.9 \\ 
        L-Ptr-Gen \citep{su-etal-2019-improving} $\dagger$ & 67.2 & 60.3 & 50.2 & 78.9 & 62.9 & 74.9 \\ 
        RUN $\dagger$ \citep{liu2020incomplete} & \textbf{70.5} & 61.2 & 49.1 & \textbf{79.1}& 61.2 & 74.7 \\ 
        AcNet & 66.7 & \textbf{61.3} & \textbf{52.7} & 76.6 & \textbf{63.5} & \textbf{77.6} \\
        \hline
    \end{tabular}
    \caption{The experimental results on CANARD dataset. $\dagger$: Results from \citealp{liu2020incomplete}.}
    \label{tab:canard}
\end{table*}

\begin{table*}[b]
    \centering
    \begin{tabular}{lccccccccc}
    \hline
        ~ & EM & Bleu-1 & Bleu-2 & Bleu-4 & ROUGE-1 & ROUGE-2 & ROUGE-L \\ \hline
        Positive & 16.1 & 63.1 & 57.1 & 48.2 & 73.9 & 59.6 & 75.0  \\ 
        Negative & 73.6 & 94.0 & 91.9 & 86.0 & 96.3 & 92.8 & 96.6  \\ \hline
    \end{tabular}
    \caption{Results of CANARD of ActNet on Positive set and Negative set.} 
    \label{tab:postiave_and_negative}
\end{table*}

In addition, Table \ref{tab:canard} reports the comparison results between several baseline models and state-of-the-art competitors. The ActNet delivers compelling results across almost all automatabtic metrics achieving a BLEU-4 score of 52.7\% and a ROUGE-L score of 77.6 on the CANARD test dataset, improving the state-of-the-art score by 2.5\% and 2.9\% respectively.

Following \citealp{su-etal-2019-improving}, we divide the CANARD test data into two categories: Positive data where rewriting is needed and Negative data, which do not need any rewriting. Furthermore, we further analyze the performance of the model on the two kinds of data. As illustrated in Table \ref{tab:postiave_and_negative}, ActNet performs better on Negative samples, suggesting that our action-based model can well identify the question that does not need rewriting. 

To further analyze the effectiveness of our method, we conduct experiments on the Chinese datasets, including REWRITE and Restoration-200K, the results of which are demonstrated in Table \ref{tab:REWRITE} and Table \ref{tab:restoration}. 

On the REWRITE test set, the ActNet achieves the copy network T-Ptr-$\lambda$ by 12.1\% and surpasses all existing state-of-the-art models with 1.6 points compared to RUN + BERT model with 66.4 points. Because of the lack of negative samples, our model is trained and estimated based on positive samples. We cannot directly compare with other models. However, we assert that better results will be obtained when negative samples are introduced based on the analysis results of Table \ref{tab:postiave_and_negative}. 

\begin{table*}[htbp]
    \centering
    \begin{tabular}{lcccccccc}
    \hline
        ~ & EM$\ast$ & BLEU-1 & BLEU-2 & BLEU-4 & ROUGE-2 & ROUGE-L  \\ \hline
        L-Gen $\dagger$ & 47.3 & ~ & 81.2 & 73.6 & 80.9 & 86.3  \\
        L-Ptr-Gen $\dagger$ & 50.5 & ~ & 82.9 & 75.4 & 83.8 & 87.8\\ 
        L-Ptr-Net$\dagger$ & 51.5 & ~ & 82.7 & 75.5 & 84.0 & 88.2  \\ 
        L-Ptr-$\lambda$ $\dagger$ & 42.3 & ~ & 82.9 & 73.8 & 81.1 & 84.1 \\ 
        T-Gen $\dagger$ & 35.4 & ~ & 72.7 & 62.5 & 74.5 & 82.9 \\ 
        T-Ptr-Gen $\dagger$ & 53.1 & ~ & 84.4 & 77.6 & 85.0 & 89.1 \\
        T-Ptr-Net $\dagger$ & 53.0 & ~ & 83.9 & 77.1 & 85.1 & 88.7 \\ 
        T-Ptr-$\lambda$ $\dagger$ & 52.6 & ~ & 85.6 & 78.1 & 85.0 & 89.0 \\ 
        RUN $\dagger$ \citep{liu2020incomplete} & \textbf{53.8} & ~ & \textbf{86.1} & 79.4 & 85.1 & 89.5 \\ 
        RAST $\mathsection$ \citep{hao2020robust} & 64.3 & 90.6 & 90.2 & \textbf{88.2} & \textbf{88.9} & \textbf{91.5}  \\ \hline
        T-Ptr-$\lambda$ + BERT $\dagger$ & 57.5 & ~ & 86.5 & 79.9 & 86.9 & 90.5 \\ 
        RUN + BERT $\dagger$ & 66.4 & ~ & 91.4 & 86.2 & \textbf{90.4} & 93.5 \\ 
        BERT+SRL \citep{xu2020semantic}  & 60.5 & 89.0 & 86.8 & 77.8 & 85.9 & 90.5 \\
        RAST+RL-BLEU $\mathsection$ & 64.4 & 89.9 & \textbf{89.6} & \textbf{87.2} & 88.7 & 91.2  \\
        RAST+RL-GPT2 $\mathsection$ & 63.0 & 89.2 & 88.8 & 86.9 & 88.2 & 90.7  \\
        ActNet & \textbf{68.0} & \textbf{91.5} & 89.5 & 85.0 & \textbf{90.4} & \textbf{93.8} \\ \hline
    \end{tabular}
    \caption{The experimental results on REWRITE dataset. $\dagger$: Results from \citealp{liu2020incomplete}. $\mathsection$: Results from \citealp{hao2020robust}. $\ast$: Note that the EM scores are all calculated on positive samples. }
    \label{tab:REWRITE}
\end{table*}

The experimental results on the test set of Restoration-200K are shown in Table \ref{tab:restoration}. We can see that the ActNet has the most winning cases (five out of eight) in the comparison of various evaluation metrics. Our proposed model achieves high numerical performance on ROUGE scores, carrying the positive performance gaps of 0.0\%(ROUGE-1), 1.1\%(ROUGE-2), and 1.0\%(ROUGE-L). The performance disadvantage of the model is exposed by BLEU scores, which shows a significant drop in the scores of three different orders. At the exact match, the strictest metric, our model got a 53.2\% EM score, outperforming its competitors by at least 3.9 percentage points.

\begin{table*}[htbp]
    \centering
    \begin{tabular}{lcccccccc}
    \hline
        ~ & EM & B1 & B2 & B3 & B4 & R1 & R2 & RL \\ \hline
        TRANS-PG+BERT \citep{hao2020robust} & 49.0 & 88.0 & 87.0 & 85.9 & 84.4 & 90.1 & 84.5 & 89.8 \\
        BERT+SRL\citep{xu2020semantic} & 49.1 & 90.6 & \textbf{89.7} & 88.6 & 87.2 & 91.1 & 85.0 & 90.0 \\
        PAC\citep{pan-etal-2019-improving} & ~ & 89.9 & 86.3 & ~ & ~ & 91.6 & 82.8 & ~ \\ 
        RUN\citep{liu2020incomplete} & 49.3 & \textbf{92.0} & 89.1 & 86.4 & 83.6 & \textbf{92.1} & 85.4 & 89.5 \\
        RAST\citep{hao2020robust} & 48.7 & 89.7 & 88.8 & 87.6 & 86.1 & 91.1 & 84.9 & 87.8 \\
        RAST+RL-BLEU & 48.8 & 90.4 & 89.6 & \textbf{88.5} & 87.0 & 91.2 & 84.8 & 87.9 \\
        RAST+RL-GPT2 & 47.8 & 89.7 & 88.9 & 87.7 & 86.2 & 90.9 & 85.0 & 87.6 \\ 
        ActNet & \textbf{53.2} & 87.1 & 85.0 & 85.2 & \textbf{88.0} & \textbf{92.1} & \textbf{86.5} & \textbf{91.0} \\ \hline
    \end{tabular}
    \caption{The experimental results on the Restoration-200K dataset.} 
    \label{tab:restoration}
\end{table*}

\section{Analysis}
We provide several common cases to illustrate that our ActNet can restore co-references and omissions. In addition, we further analyze the shortcomings of this method through several bad cases.

\begin{figure}[htbp]
    \centering
    \includegraphics[width=\linewidth]{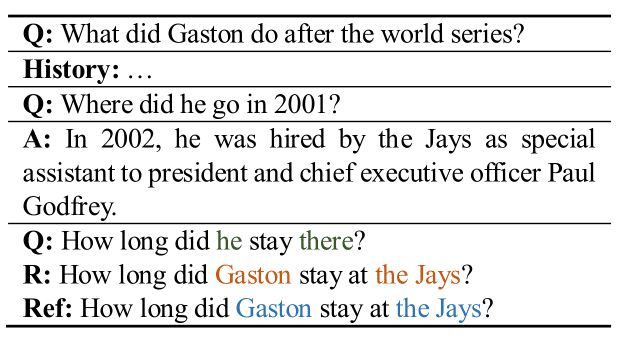}
    \caption{A successful case of ActNet. “R” means the prediction of  ActNet and “Ref” means the ground truth. The different colors indicate the differences between the original question, reference and prediction.}
    \label{fig:fig4}
\end{figure}

\paragraph{Strengths:} Our method is good at dealing with situations where omitted or referred information in the question has a continued counterpart in the conversation history. As illustrated in Figure \ref{fig:fig4}, the case where demonstrative pronoun refers to a continued segment of the context, such as “he” points to “Gaston” and “there” points to “the Jays”, is relatively easy for ActNet to handle. Besides, ActNet can find it from conversation history if the omission information contained in the question is a continued string, such as “Arizona State University” omitted after “graduate” described in Figure \ref{fig:fig1}.

\begin{figure}[htbp]
    \centering
    \includegraphics[width=\linewidth]{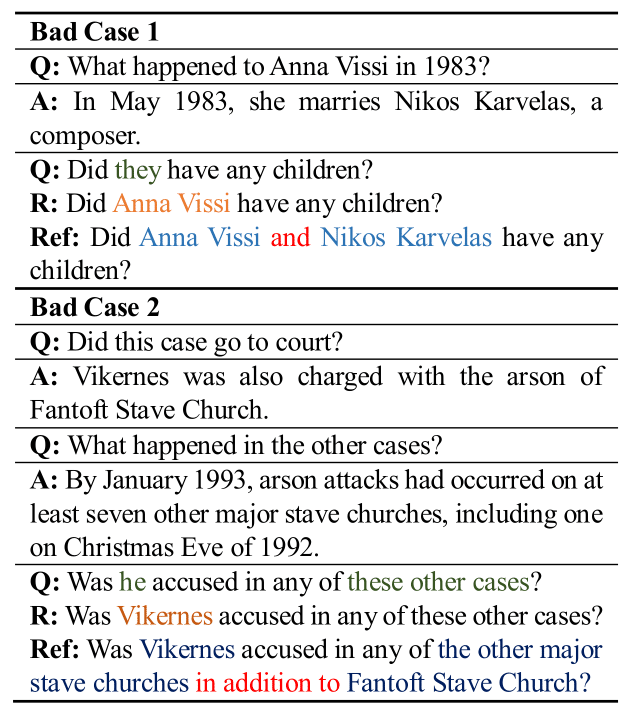}
    \caption{Two bad cases of ActNet.}
    \label{fig:fig5}
\end{figure}

\paragraph{Weaknesses:} However, it would be very tricky for ActNet if a candidate clue refers to multi-information in the conversation history. For example, “they” refers to the disordered combination of “Anna Vissi” and “Nikos Karvelas” in bad case 1 of Figure \ref{fig:fig5}. Especially, ActNet cannot reorganize all candidates retrieved from conversation history according to syntactic rules. Take bad case 2 for an example,  “these other cases” refers to the combination of “the other major stave churches” and “Fantoft Stave Church”, where all components are required to be combined reasonably.

\section{Conclusion}
This paper proposes a conversation question rewrite network, which first employs pre-trained language model BERT-base as the encoder module to model context interaction. Then add a detecting module to find out the candidates of co-reference or ellipsis in the question while assigning the corresponding action to each candidate span. It uses the comprehension module to search for matching phrases from the conversation context based on the candidate clues. In the experiments, we show that our design is significantly and consistently comparable to state-of-the-art comparison networks. In the future, we will explore the problem that a candidate clue of the question refers to multi-information of conversation history. Beyond that, more detailed ablation experiments could be conducted to explore the performance of the network structure on two sub-tasks: Co-reference Resolution and Information Completion. 

\bibliography{anthology,acl2020}
\bibliographystyle{acl_natbib}

\appendix

\section{Appendices}
\label{sec:appendix}
\subsection{Stop words}
Stop words include “of”, “and”, “in”, “to”, “as”, “for”, 'on', “with”, “by”, “at”, “from”, “but”, “or”,  “up”, “out”, “after”, “into”, “about”, “over”, “then”, “some”, “little”, “just”, “than”, “around”, “both”, “off”, “until”, “any”, “including”, “away”, “the”, “addition”, “’s” and “?”.

\end{document}